\documentclass{article}

\usepackage{microtype}
\usepackage{graphicx}
\usepackage{subcaption}
\usepackage{booktabs} 
\usepackage{multirow}

\usepackage{hyperref}

\usepackage[preprint]{icml2026}

\usepackage{amsmath}
\usepackage{amssymb}
\usepackage{mathtools}
\usepackage{amsthm}

\usepackage[capitalize,noabbrev]{cleveref}

\theoremstyle{plain}

\theoremstyle{definition}

\theoremstyle{remark}

\usepackage[textsize=tiny]{todonotes}

\icmltitlerunning{Non-linear Interventions on Large Language Models}

\begin{document}

\twocolumn[
  \icmltitle{Non-linear Interventions on Large Language Models}



  \icmlsetsymbol{equal}{*}

  \begin{icmlauthorlist}
    \icmlauthor{Sangwoo Kim}{snu}
  \end{icmlauthorlist}

  \icmlaffiliation{snu}{Department of Linguistics, Seoul National University, Republic of Korea}
  \icmlcorrespondingauthor{Sangwoo Kim}{hemy0101@snu.ac.kr}
  \icmlkeywords{Machine Learning, ICML}

  \vskip 0.3in
]



\printAffiliationsAndNotice{}  

\begin{abstract}
  Intervention is one of the most representative and widely used methods for understanding the internal representations of large language models (LLMs). However, existing intervention methods are confined to linear interventions grounded in the Linear Representation Hypothesis, leaving features encoded along non-linear manifolds beyond their reach. In this work, we introduce a general formulation of intervention that extends naturally to non-linearly represented features, together with a learning procedure that further enables intervention on \emph{implicit} features lacking a direct output signature. We validate our framework on refusal bypass steering, where it steers the model more precisely than linear baselines by intervening on a non-linear feature governing refusal.
\end{abstract}

\section{Introduction}

Intervention is a central tool in understanding inner representations of large language models (LLMs). By modifying a model's internal activations during inference and observing the resulting change in output, interventions yield causal evidence linking specific components to model behavior~\cite{NEURIPS2020_92650b2e, meng2022locating, geiger2024findingalignmentsinterpretablecausal}. Through intervention, we can understand how interpretable features are represented in the model's hidden states~\cite{huang-etal-2024-ravel}, and by doing so, effectively control via steering LLM behaviors that are hard to address through prompting alone, such as response style~\cite{turner2025steering}, hallucination~\cite{li2023inferencetime}, and refusal~\cite{arditi2024refusallanguagemodelsmediated}.

Most existing methods for intervening on such features are linear: they modify hidden states by adding or ablating a fixed direction in activation space~\cite{li2023inferencetime, arditi2024refusallanguagemodelsmediated}. This design rests on the Linear Representation Hypothesis (LRH), which posits that interpretable concepts are encoded as directions in the model's representation space~\cite{mikolov-etal-2013-linguistic, park2023the}. Recent work, however, has shown that some concepts are instead represented along non-linear manifolds---for example, days of the week organized as a circular structure~\cite{engels2025not}, and the geometric structures used to perform counting~\cite{gurnee2025when}. Since linear interventions can by construction only manipulate features encoded as directions, they cannot reach this class of non-linearly represented structure.

To overcome this limitation, we introduce a general formulation of non-linear intervention. We generalize linear interventions by replacing their underlying linear transformation with an invertible non-linear feature map. In addition, we propose a procedure for learning this map via \emph{interchange interventions}~\cite{geiger2024findingalignmentsinterpretablecausal}, together with a loss design that extends the procedure to \emph{implicit} features. 

We instantiate the framework on refusal bypass steering~\cite{arditi2024refusallanguagemodelsmediated, wollschlager2025the}, a representative implicit-feature intervention task. Our non-linear intervention attains steering effectiveness comparable to strong linear baselines while editing activations at orders of magnitude fewer hidden-state locations. Further analysis indicates that this advantage stems from a genuinely non-linear feature map governing refusal that resides primarily at the model's middle layers.

Our work makes three contributions. First, we propose a general formulation of non-linear intervention that subsumes existing linear interventions and naturally extends to features encoded along non-linear manifolds. Second, we introduce a learning procedure for the non-linear feature map based on interchange interventions, together with a loss design that further enables learning \emph{implicit} features that do not surface in the model's output. Third, we empirically apply our framework to refusal bypass steering and show that intervening on a non-linear feature governing refusal steers the model more precisely than linear baselines. Code is available at \url{https://anonymous.4open.science/r/nonlinear-intervention-77AC/}.


\section{A General Formulation of Non-linear Interventions}
\label{sec:method}

\subsection{Linear Interventions as Change of Basis}
\label{sec:linear-interventions}

A common approach to intervene on a language model $\mathcal{M}$ at inference time is to identify \emph{linear feature directions} in its representation space and perturb the hidden state along them. Given orthonormal feature directions $\{\mathbf{v}_{i}\}_{i=1}^{k} \subset \mathbb{R}^{d}$ and corresponding scalar coefficients $\{\alpha_{i}\}_{i=1}^{k}$, this linear intervention modifies the hidden state $\mathbf{h} \in \mathbb{R}^{d}$ at a chosen position as
\begin{equation}
    \mathbf{h} \;\leftarrow\; \mathbf{h} + \sum_{i=1}^{k} \alpha_{i} \mathbf{v}_{i}.
    \label{eq:linear-multi}
\end{equation}

Eq.~\eqref{eq:linear-multi} can equivalently be viewed as a change of basis. Let $W \in \mathbb{R}^{d \times d}$ be an orthogonal matrix that maps the representation space onto a \emph{linear feature space}, satisfying $W^{\top} \mathbf{e}_{i} = \mathbf{v}_{i}$, where $\{\mathbf{e}_{i}\}_{i=1}^{d}$ is the standard basis of $\mathbb{R}^{d}$. Each coordinate of $W \mathbf{h}$ is then the activation value of one feature at $\mathbf{h}$, and Eq.~\eqref{eq:linear-multi} can be rewritten as
\begin{equation}
    \mathbf{h} \;\leftarrow\; W^{-1}\!\left( W \mathbf{h} + \sum_{i=1}^{k} \alpha_{i} \mathbf{e}_{i} \right),
    \label{eq:linear-W}
\end{equation}
that is, $W$ maps $\mathbf{h}$ into the feature space, the intervention perturbs the resulting feature coordinates along their axes, and $W^{-1}$ maps back to the original representation space.

\subsection{Non-linear Interventions via Invertible Feature Maps}
\label{sec:non-linear-interventions}

Linear interventions can only manipulate features that admit a linear encoding, yet not all interpretable features in $\mathcal{M}$'s representation space lie along linear directions~\cite{engels2025not, kantamneni2025language}. To accommodate non-linear features, we replace $W$ with an invertible \emph{non-linear feature map} $f_{\theta} : \mathbb{R}^{d} \to \mathbb{R}^{d}$ parameterized by $\theta$. By direct analogy with Eq.~\eqref{eq:linear-W}, the non-linear intervention takes the form
\begin{equation}
    \mathbf{h} \;\leftarrow\; f_{\theta}^{-1}\!\left( f_{\theta}(\mathbf{h}) + \sum_{i=1}^{k} \alpha_{i} \mathbf{e}_{i} \right).
    \label{eq:non-linear-intervention}
\end{equation}
$f_{\theta}$ maps $\mathbf{h}$ to its feature-space coordinates; the intervention perturbs these coordinates along their axes by the coefficients $\{\alpha_{i}\}$; and $f_{\theta}^{-1}$ maps the modified features back into the original hidden-state space. Eq.~\eqref{eq:linear-W} is recovered as the special case in which $f_{\theta}$ is restricted to a linear map.

\section{Learning $f_{\theta}$ via Interchange Intervention}
\label{sec:training}

\subsection{Training Objective via Interchange Intervention}

This section describes how we learn the non-linear feature map $f_{\theta}$ associated with a target feature $\mathcal{F}$. We train $f_{\theta}$ via \emph{interchange interventions}~\cite{geiger2024findingalignmentsinterpretablecausal}, which provide causal supervision over $\mathcal{F}$ by transferring its value between hidden states drawn from contrasting inputs.
We first prepare two sets of prompts: a positive set $\mathcal{D}^{+} = \{x_i^{+}\}_{i=1}^{N}$ of inputs that exhibit the feature $\mathcal{F}$, and a negative set $\mathcal{D}^{-} = \{x_i^{-}\}_{i=1}^{N}$ of inputs that do not. For each pair $(x^{-}, x^{+}) \in \mathcal{D}^{-} \times \mathcal{D}^{+}$, we forward both inputs through $\mathcal{M}$ up to the intervention site, obtaining hidden states $\mathbf{h}^{-}, \mathbf{h}^{+} \in \mathbb{R}^{d}$. The \emph{interchange intervention} replaces the targeted coordinates of $f_{\theta}(\mathbf{h}^{-})$ with those of $f_{\theta}(\mathbf{h}^{+})$:
\begin{equation}
    \mathbf{h}^{-} \;\leftarrow\; f_{\theta}^{-1}\!\left( f_{\theta}(\mathbf{h}^{-}) + \sum_{i=1}^{k} \alpha_{i} \mathbf{e}_{i} \right),
    \label{eq:interchange}
\end{equation}
where $\alpha_{i} = \big( f_{\theta}(\mathbf{h}^{+}) - f_{\theta}(\mathbf{h}^{-}) \big)_{i}$. This recovers Eq.~\eqref{eq:non-linear-intervention} with the coefficients pinned to $x^{+}$. Let $\mathcal{M}_{\text{int}}(x^{-}, x^{+};\, \theta)$ denote the intervened forward pass of $\mathcal{M}$. With pairs sampled as $(x^{-}, x^{+}) \sim \mathcal{D}^{-} \times \mathcal{D}^{+}$ and $\mathcal{M}$ frozen, we train $\theta$ to minimize
\begin{equation}
    \mathcal{L}(\theta) \;=\; \mathbb{E}_{(x^{-},\, x^{+})} \!\left[\, \ell\!\left( \mathcal{M}_{\text{int}}(x^{-}, x^{+};\, \theta) \right) \,\right],
    \label{eq:loss}
\end{equation}
where $\ell$ is designed so that its minimization makes $\mathcal{M}_{\text{int}}$ exhibit $\mathcal{F}$.

\subsection{Loss Design for Implicit Features}

When $\mathcal{F}$ has a direct output signature, $\ell$ can be defined on $\mathcal{M}$'s output distribution to encourage or suppress specific tokens. For implicit features that do not surface in the output, such as refusal or stylistic shifts, no such direct objective for training $f_\theta$ is available, making the loss design non-trivial. We propose a self-supervised-style loss that learns $f_\theta$ from data alone, by enforcing causal influence over many features correlated with $\mathcal{F}$. 

For each (layer, token position) site $s$ lying causally downstream of the intervention site, we extract a feature direction $v_{s} \in \mathbb{R}^{d}$ as the class-mean difference of unintervened activations $h_{s}(x)$ over $\mathcal{D}^{+}$ versus $\mathcal{D}^{-}$. Projections onto $v_{s}$ are therefore correlated with $\mathcal{F}$. We keep only those sites at which $v_{s}^{\top} h_{s}(x)$ separates $x^{+}$ from $x^{-}$ with AUC above a threshold $\tau$, collected into $\mathcal{S}$. Both $v_{s}$ and $\mathcal{S}$ are computed once from $\mathcal{D}^{\pm}$ before training and held fixed.

Letting $h_{s}^{\text{int}}$ denote the hidden state at $s$ under the intervened forward pass and $\mu_{s}^{+} = \mathbb{E}_{x \sim \mathcal{D}^{+}}[\, v_{s}^{\top} h_{s}(x) \,]$ the mean projection of $x^{+}$, we take $\ell$ in Eq.~\eqref{eq:loss} to be
\begin{equation}
    \ell(\mathcal{M}_{\text{int}}) \;=\; \sum_{s \in \mathcal{S}} \max\!\big(\, 0,\;\; \mu_{s}^{+} - v_{s}^{\top} h_{s}^{\text{int}} \,\big).
    \label{eq:internal-loss}
\end{equation}
The hinge form saturates once a site's projection reaches $\mu_{s}^{+}$, preventing overfitting to any single site. Minimizing $\mathcal{L}$ thus drives $f_{\theta}$ to discover a feature whose intervention causally aligns many $\mathcal{F}$-correlated components at once.

This particular instantiation is one natural choice. The direction extractor, site-selection criterion, and surrogate loss are not fixed; each can be substituted within the framework of Eq.~\eqref{eq:loss}.

\section{Experiments}
\label{sec:experiments}

\subsection{Setup}
\label{sec:experiment-setup}

We evaluate non-linear intervention on refusal bypass steering in safety-aligned LLMs, a representative steering task, to empirically validate our proposed non-linear intervention framework.
We use this setting as an implicit-feature intervention task: the
target feature $\mathcal{F}$ is \emph{bypass refusal}, which is not explicitly manifested in any specific token output of the model.

\paragraph{Models and data.}
We evaluate all methods on Llama-3-8B-Instruct~\cite{grattafiori2024llama3herdmodels}
and Qwen2.5-7B-Instruct~\cite{qwen2025qwen25technicalreport}, with model weights frozen throughout.
We construct $\mathcal{D}^{+}$ from $2{,}000$ harmless Alpaca prompts~\cite{alpaca}
that elicit compliant responses from the model, and $\mathcal{D}^{-}$ from
$2{,}000$ harmful SALAD-Bench prompts~\cite{li2024salad} that elicit refusals.

\paragraph{Evaluation.}
We quantitatively evaluate intervention quality along two axes: \emph{how
strongly} the intervened model exhibits the target feature $\mathcal{F}$,
and \emph{how much} we perturbed the model to achieve it.
To measure whether the intervened model produces genuinely meaningful
responses to harmful prompts rather than merely avoiding surface-level
refusals, we use the StrongREJECT score~\citep{souly2024strongreject}. StrongREJECT
returns a score in $[0,1]$ over $313$ harmful prompts, computed by a
LLM-based evaluator. To measure how much we intervened, we record the total intervention
magnitude applied to the model. Specifically, for every edited site, we
compute the $\ell_2$ distance between the pre-intervention hidden state
$h$ and the post-intervention hidden state $h'$, and sum them:
\[
  \frac{1}{|\mathcal{D}_{\mathrm{test}}|}
  \sum_{x \in \mathcal{D}_{\mathrm{test}}}
  \sum_{s \in \mathcal{E}_x}
  \| h'_{s}(x) - h_{s}(x) \|_{2},
\]
where $\mathcal{E}_x$ is the set of sites the method edits
when generating a response to prompt $x$.

\paragraph{Baselines.}
As baselines, we compare against two representative linear intervention
methods for refusal steering: \emph{Difference In Means} (DIM)~\cite{arditi2024refusallanguagemodelsmediated}
and \emph{Refusal Direction Optimization} (RDO)~\cite{wollschlager2025the}. DIM extracts a refusal direction from the class-mean activation difference. RDO learns a refusal direction with the ablation, addition, and retain
losses. Both methods can be evaluated under two intervention schemes: \emph{ablation}, which projects out the refusal direction at every token and every module output, and \emph{actadd}, which adds the direction scaled by a fixed coefficient $\alpha$ at a designated layer for every token.

\paragraph{Our method.}
We instantiate $f_{\theta}$ as an i-ResNet~\cite{pmlr-v97-behrmann19a}, an invertible non-linear neural network. We use $k=1$ for direct comparison with one-dimensional linear baselines. We heuristically select one intervention position per model.
At inference time, to avoid requiring a harmless prompt $x^+$ for each query, we precompute the mean feature activation over a set of harmless prompts $x^+$ and use this cached value.
Specifically, we compute the mean
first-coordinate value of $f_{\theta}(h)$ over the harmless training set at the selected site, denoted $\bar{\mu}^{+}$, and clamp each test
activation to this value.

\subsection{Results}

\begin{figure}[t]
\centering
\includegraphics[width=0.95\columnwidth]{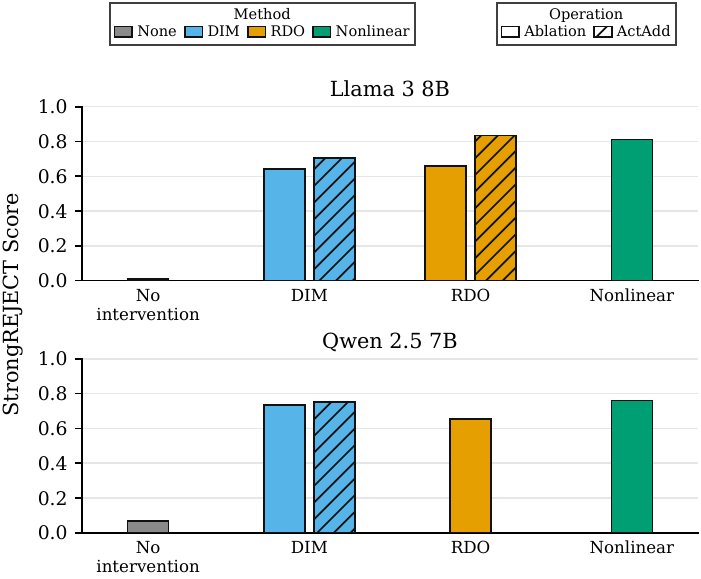}
\caption{StrongREJECT scores on Llama 3 8B and Qwen 2.5 7B without intervention, with the linear baselines, and with our non-linear intervention.\protect\footnotemark}
\label{fig:sr_scores}
\end{figure}
\footnotetext{Although we faithfully reproduced the official RDO repository, the RDO ActAdd variant on Qwen yielded a StrongREJECT score of 0.020 in our runs. We judged this result to be unreliable and therefore excluded it from the figure and table.~\citealt{wollschlager2025the} reports ActAdd performance comparable to that of the Ablation variant.}

Figure~\ref{fig:sr_scores} reports StrongREJECT scores under no intervention, the linear baselines, and our non-linear intervention. Although the non-linear method intervenes at a single position per sample, it reaches scores comparable to the baselines, which instead modify activations at a much larger number of sites.

\begin{table}[t]
\centering
\small
\caption{Per-sample intervention magnitude. \emph{Sites}: avg.\ positions edited per example; \emph{$L_2$}: total edit norm. Our non-linear method edits one site, with an $L_2$ norm over two orders of magnitude below every linear baseline.}
\label{tab:intervention_l2}
\begin{tabular}{llrr}
\toprule
Model & Method & Sites & $L_2$ \\
\midrule
\multirow{5}{*}{Llama 3 8B}
 & DIM (Ablation) & 54{,}778 & 2{,}898.7 \\
 & DIM (ActAdd)   &     572  & 1{,}816.2 \\
 & RDO (Ablation) & 59{,}396 & 1{,}812.1 \\
 & RDO (ActAdd)   &     619  & 1{,}964.5 \\
 & \textbf{non-linear} & \textbf{1} & \textbf{3.5} \\
\midrule
\multirow{4}{*}{Qwen 2.5 7B}
 & DIM (Ablation) & 49{,}661 & 41{,}376.2 \\
 & DIM (ActAdd)   &     591  & 24{,}676.3 \\
 & RDO (Ablation) & 53{,}568 & 15{,}474.0 \\
 & \textbf{non-linear} & \textbf{1} & \textbf{28.3} \\
\bottomrule
\end{tabular}
\end{table}

As shown in Table~\ref{tab:intervention_l2}, the total magnitude of the non-linear intervention is more than two orders of magnitude smaller than that of every linear baseline. This empirically suggests that our non-linear intervention captures the feature inside the model more precisely than the linear baselines do.

\section{Analysis}

\subsection{Linear Intervention at the Loss Sites}
\label{sec:analysis-linear-loss-sites}

Our objective in Eq.~\eqref{eq:internal-loss} pushes $v_{s}^{\top} h_{s}^{\text{int}}$ toward $\mu_{s}^{+}$ at every site $s \in \mathcal{S}$. We ask whether the gain over linear baselines comes solely from our choice of $\mathcal{S}$ and the directions $\{v_{s}\}$. Concretely, at every $s \in \mathcal{S}$ we apply a linear intervention along $v_{s}$ with a coefficient chosen so that $v_{s}^{\top} h_{s}$ is shifted to $\mu_{s}^{+}$. As shown in Table~\ref{tab:linear-loss-sites}, this falls substantially short of our method on both models, indicating that $f_{\theta}$ does more than align these projections: it implements a non-linear, causal manipulation of the refusal feature that \emph{induces} the per-site alignment as a downstream effect.

\begin{table}[t]
\centering
\small
\caption{StrongREJECT after intervening linearly at every loss site $s \in \mathcal{S}$ along $v_{s}$, versus our learned non-linear intervention at a single site.}
\label{tab:linear-loss-sites}
\begin{tabular}{lrr}
\toprule
Method & Llama 3 8B & Qwen 2.5 7B \\
\midrule
Linear at loss sites & 0.689 & 0.415 \\
\textbf{Non-linear (ours)} & \textbf{0.813} & \textbf{0.760} \\
\bottomrule
\end{tabular}
\end{table}

\subsection{Role of Non-linearity}
\label{sec:analysis-linear-ftheta}

To examine the role of $f_{\theta}$'s non-linearity, we re-train it under the same loss and pipeline as Section~\ref{sec:training}, but constrained to a linear map. As reported in Table~\ref{tab:linear-ftheta}, StrongREJECT drops sharply on both models, suggesting that the non-linearity of $f_{\theta}$ clearly contributes to the effectiveness of our intervention.

\begin{table}[t]
\centering
\small
\caption{StrongREJECT when $f_{\theta}$ is constrained to a linear map, versus our i-ResNet $f_{\theta}$, trained under the same objective.}
\label{tab:linear-ftheta}
\begin{tabular}{lrr}
\toprule
Method & Llama 3 8B & Qwen 2.5 7B \\
\midrule
Linear $f_{\theta}$ & 0.010 & 0.072 \\
\textbf{Non-linear $f_{\theta}$ (ours)} & \textbf{0.813} & \textbf{0.760} \\
\bottomrule
\end{tabular}
\end{table}

\subsection{Layer Sweep}
\label{sec:analysis-layer-sweep}

\begin{figure}[t]
\centering
\includegraphics[width=0.95\columnwidth]{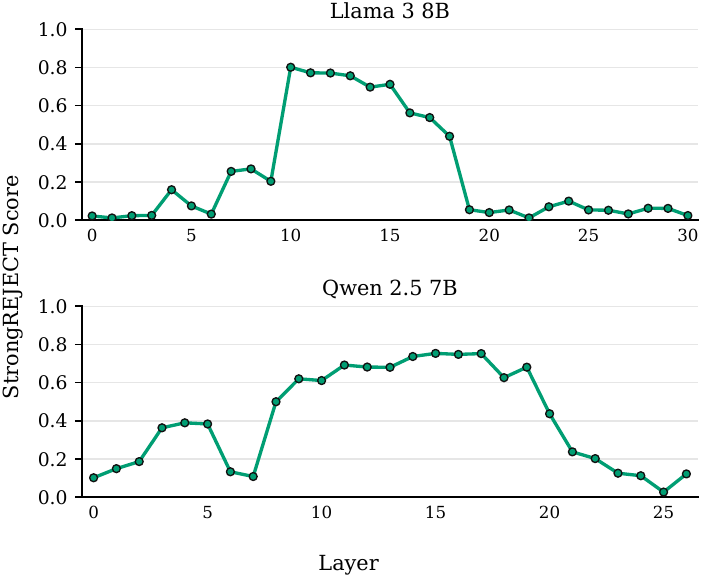}
\caption{StrongREJECT across intervention layers on Llama 3 8B and Qwen 2.5 7B with our non-linear intervention.}
\label{fig:layer-sweep}
\end{figure}

To check whether our non-linear intervention is effective at any layer, we repeat the same training and evaluation under the same setup while varying only the intervention layer. Figure~\ref{fig:layer-sweep} shows that, for both models, the intervention is largely ineffective at early layers, peaks at middle layers, and declines at later layers. This suggests that our method is not so powerful that it finds an effective feature map at any layer; rather, it captures a non-linear feature governing refusal/compliance that resides primarily in the middle layers.

\section{Conclusion}
\label{sec:conclusion}

We overcome the linearity restriction of existing interventions by providing a general formulation of non-linear intervention. Beyond the formulation itself, we propose a procedure for learning the non-linear feature map via interchange interventions, together with the idea of supervising this procedure for \emph{implicit} features through hidden states of the model that are correlated with the target feature. We instantiate this framework on bypass refusal steering, a representative implicit-feature steering task; our experiments and analysis show that non-linear intervention effectively steers refusal behavior and that, in doing so, it uncovers a non-linear feature inside the model.

Several limitations remain, however. We realize $f_{\theta}$ only as an i-ResNet, and the same framework admits other invertible architectures. We also do not overcome a limitation shared
with linear interventions: the position at which to intervene
must still be chosen heuristically. We leave both directions
to future work.

\nocite{langley00}

\bibliography{example_paper}
\bibliographystyle{icml2026}

\newpage
\appendix
\onecolumn

\section{Additional Experimental Details}
\label{app:experimental-details}

\subsection{i-ResNet Feature Map}
\label{app:iresnet-details}

We instantiate the invertible feature map $f_\theta : \mathbb{R}^d \to \mathbb{R}^d$
as a composition of $M = 2$ invertible residual blocks,
\[
    f_\theta \;=\; \phi_{\theta_M} \circ \cdots \circ \phi_{\theta_1},
    \qquad
    \phi_{\theta_m}(z) \;=\; z + g_{\theta_m}(z).
\]
The residual branch $g_{\theta_m} : \mathbb{R}^d \to \mathbb{R}^d$ is a
two-layer MLP with hidden width $128$ and a LeakyReLU nonlinearity (negative
slope $0.1$). To guarantee invertibility via $\mathrm{Lip}(g_{\theta_m}) < 1$,
we follow \citet{pmlr-v97-behrmann19a} and combine two mechanisms:
(i) every linear layer $W$ inside $g_{\theta_m}$ is replaced at every forward
pass by $\widetilde{W} = W \cdot \min(1, 1/\hat\sigma(W))$, where
$\hat\sigma(W)$ is a running estimate of the spectral norm of $W$ obtained by
a single power iteration with a persistent left singular vector
\citep{miyato2018spectralnormalizationgenerativeadversarial}; (ii) the output of $g_{\theta_m}$ is rescaled by a
Lipschitz coefficient $\kappa = 0.6$, yielding
$\mathrm{Lip}(g_{\theta_m}) \le \kappa < 1$. We invert each block
by the contractive fixed-point iteration $x \leftarrow y - g_{\theta_m}(x)$ run
for up to $30$ steps with relative-residual tolerance $10^{-5}$, and propagate
gradients through the inverse via implicit differentiation.

\subsection{Intervention Sites}
\label{app:intervention-sites}

All layer indices are zero-indexed transformer block indices. Our non-linear
intervention is placed at the \texttt{block\_output} representation, i.e., the
residual-stream state after the selected transformer block. Token positions are
negative indices into the fully formatted and tokenized chat prompt after the
generation prompt has been appended; position $-1$ denotes the final prompt
token immediately before generation starts.

For the main non-linear experiments, we intervene on exactly one site per
example. The sites are:
\begin{table}[h]
\centering
\small
\caption{Main non-linear intervention sites. Token positions are counted from
the end of the formatted prompt.}
\label{tab:appendix-nonlinear-sites}
\begin{tabular}{lccc}
\toprule
Model & Layer & Token position & Evaluation target \\
\midrule
Llama 3 8B & 10 & $-3$ & harmless mapped-coordinate mean \\
Qwen 2.5 7B & 14 & $-4$ & harmless mapped-coordinate mean \\
\bottomrule
\end{tabular}
\end{table}

\end{document}